\pdfoutput=1

\documentclass[11pt]{article}

\usepackage[]{EMNLP2022}

\usepackage{times}
\usepackage{latexsym}

\usepackage[T1]{fontenc}

\usepackage[utf8]{inputenc}

\usepackage{microtype}

\usepackage{enumitem}
\setlist{nosep}
\usepackage{graphicx}

\def\elevant{Elevant}
\newcommand{\entity}[2]{\textit{#1 (#2)}}
\newcommand{\mention}[1]{\underline{#1}}

\title{ELEVANT: A Fully Automatic Fine-Grained\\Entity Linking Evaluation and Analysis Tool}

\author{
Hannah Bast \and Matthias Hertel \and Natalie Prange \\
University of Freiburg \\
Freiburg, Germany \\
\texttt{\{bast,hertelm,prange\}@cs.uni-freiburg.de}
}

\makeatletter
\def\endthebibliography{%
  \def\@noitemerr{\@latex@warning{Empty `thebibliography' environment}}%
  \endlist
}
\makeatother

\begin{document}
\maketitle
\begin{abstract}
We present \emph{\elevant{}}, a tool for the fully automatic fine-grained evaluation of a set of entity linkers on a set of benchmarks.
\elevant{} provides an automatic breakdown of the performance by various error categories and by entity type.
\elevant{} also provides a rich and compact, yet very intuitive and self-explanatory visualization of the results of a linker on a benchmark in comparison to the ground truth.
A live demo, the link to the complete code base on GitHub and a link to a demo video are provided under \url{https://elevant.cs.uni-freiburg.de} .
\end{abstract}

\section{Introduction}

Entity linking is a fundamental problem, and a first step for or component of many NLP applications.
In this paper, we consider end-to-end entity linking systems, which do the following: given a text and a set of entities,
identify each mention of one of those entities in the text and say which of these entities it is.

Due to its fundamental importance and wide applicability, there is vast literature on entity linking, and also a large number of concrete software tools.
Many publications also come with an evaluation, which compares the entity linker introduced in the publication with existing linkers, usually on a variety of benchmarks. There are also several standard benchmarks, which are found in many evaluations.
This is a positive and pleasing development.

The typical statistics include overall precision and recall, that is, which percentage of the found mentions were correct and which percentage of the correct mentions were found. It is a frequent experience that the numbers in the evaluation are very good, yet the experience when applying that entity linker in an own application are less convincing. And not so rarely, there is even trouble reproducing the results from the publication.

The problem is that overall precision, recall, and F1 tell us little about the particular strengths and weaknesses of a particular entity linker for a particular application.
Particular benchmarks often require very particular skills from an entity linker, and an entity linker may be deliberately or unknowingly tuned towards these particularities.
To find out about the strengths and weaknesses of an entity linker, one needs to look at the results in more detail, which typically has \emph{three} aspects:\\[1mm]
(1) look at particular types of errors,\\
(2) look at particular types of entities,\\
(3) look at particular pieces of text.\\[1mm]
Doing this on the raw input and output files is tedious, so that often small scripts are written to aid this process.
However, these scripts are usually quite basic and imperfect.
It also means that researchers do the same work over and over again.

It is the purpose of this paper to provide a comprehensive and easy-to-use tool,
which every entity-linking researcher can and wants to use
to analyze and understand the performance of a particular entity linker in detail.

\subsection{Contributions}

We consider these as our main contributions:
\vspace{1mm}

\begin{itemize}[parsep=1.0mm,leftmargin=0mm,itemindent=4mm,topsep=0pt]

\item We provide \emph{\elevant{}}, a tool for the fully automatic fine-grained evaluation of a given set of entity linkers on a given set of benchmarks. The evaluation has two parts: a table with overall and fine-grained statistics, and a panel that provides a rich visualization of the concrete results that contribute to a selected statistics from the table.

\item The table provides one row per experiment (a particular entity linker evaluated on a particular benchmark). Each column stands for one of a rich set of error categories: all errors, various kinds of entity detection errors, various kinds of entity disambiguation errors, errors on entities of particular type. There are controls to show and hide individual columns or groups of columns. A screenshot of this part of the tool is shown in Figure~\ref{fig:evaluation_results}.

\begin{figure*}[t]
	\resizebox{\textwidth}{!}{
		\centering
		\includegraphics{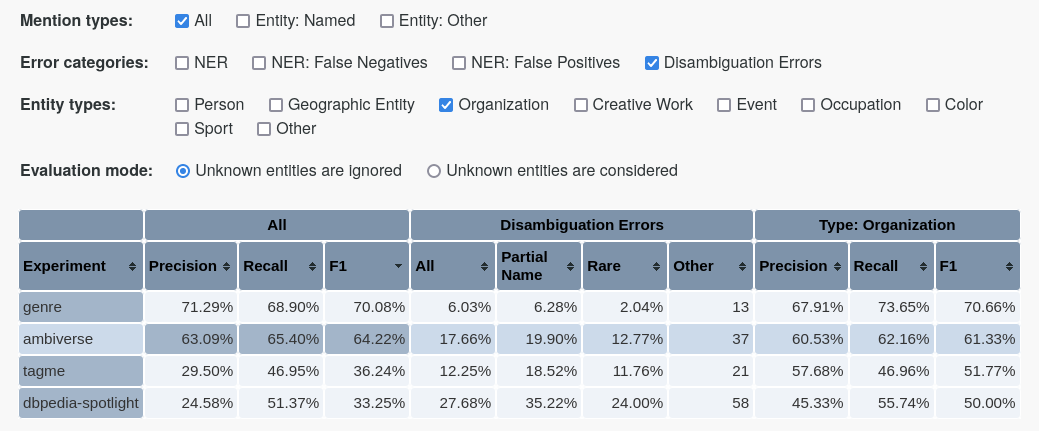}
	}
	\caption{Evaluation results for various experiments. The types used in the per-type evaluation are configurable. The real web app contains many more error categories (see Section~\ref{features:error-types}).}
	\label{fig:evaluation_results}
\end{figure*}

\item For each table cell, \elevant{} provides a rich visualization of the result of that particular entity linker on that benchmark for that category. The visualization is compact and intuitive, providing for each text record all information about true positives, false positives, and false negatives. The information is displayed with intuitive highlights and more detailed information provided on mouseover. Figure~\ref{fig:annotations} shows a screenshot of this visualization.

\begin{figure*}[t]
	\resizebox{\textwidth}{!}{
		\centering
		\includegraphics{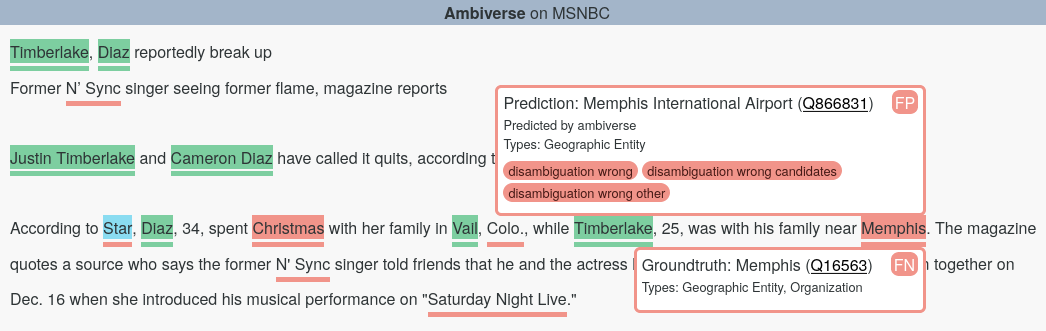}
	}
	\caption{Visualization of predicted entities (highlighted text) and ground truth labels (underlined text) for a specific system (Ambiverse) and benchmark (MSNBC). True positives are shown in green, false positives and false negatives in red and unevaluated unknown entities in blue. Detailed information for each annotation is shown on mouseover.}
	\label{fig:annotations}
\end{figure*}

\item As a result of the compact single-column visualization, \elevant{} can also provide an intuitive side-by-side comparison of the concrete results of two experiments.

\item The code for \elevant{} is open source, well documented, and easy to use. We support various standard formats for both the benchmarks and the results from the entity linkers. There is a demo website available under \url{https://elevant.cs.uni-freiburg.de},
which shows the results of a variety of well-known entity linkers evaluated on a variety of well-known benchmarks.

\end{itemize}

\section{Related Work}

GERBIL \cite{gerbil} is similar to Elevant in that it provides a web-based platform for the comparison
of a given set of entity linking systems on a given set of benchmarks.
GERBIL is widely used and has helped standardizing benchmarks and the evaluation measures on these benchmarks.
Elevant goes one step further by not only providing aggregate measures for the whole benchmarks (like precision, recall, and F1),
but a detailed breakdown of the results by error category and entity type,
and the ability to look at the results of different methods in detail,
both in comparison to the ground truth and in comparison with each other.

ORBIS \cite{orbis} is similar to Elevant in that it provides overall statistics and a visualization of individual annotations.
However, ORBIS does not provide a fine-grained error analysis and the visualization is less rich and less compact compared to that of {\elevant}.
In particular, the visualization uses two columns:
one for the annotations from the entity linker and one for those from the ground truth.
This makes it difficult to grasp the most important information at one glance and there is no support for the comparison of two entity linkers.
While the source code is publicly available and easy to install, errors can occur during usage due to dependency issues.
\elevant{} avoids this issue by providing an easy-to-use docker setup.

VEX \cite{vex} is a web app for visual error analysis of entity linking systems. Benchmark texts are displayed with highlighted predicted entities and ground truth entities. The highlights are color-coded such that true positives, false positives and false negatives can be distinguished.
VEX focuses on showing clusters of entities, that is, indicating which predicted mentions and ground truth mentions have been linked to the same entity. For this purpose, identical entities are connected via lines.
VEX does not display a system's evaluation results and does not allow direct comparison of different systems.

As part of their work on entity linking on Wikipedia, \citet{hedwig} provide a system for visualizing annotations in Wikipedia articles such as hyperlinks or an entity linking tool's entity predictions. Identical entities are visualized by using the same annotation color. The tool is not meant for evaluating linking results against a ground truth. That is, no ground truth entities are displayed and the tool does not provide information about true positives, false positives or false negatives.

\citet{wexea} provide a similar but more rudimentary system as part of their work on entity linking on Wikipedia. Predicted links are shown as hyperlinks to Wikipedia articles which correspond to the predicted entity.
The color of the hyperlink indicates whether the hyperlink is an original intra-Wikipedia hyperlink or has been added by the entity linking system. An additional color is used for predicted unknown entities.

Multiple publications propose a fine-grained evaluation of entity linking systems on different entity types or frequent linking errors.
\citet{figer} and \citet{fine-grained-types} assign fine-grained types to recognized entities.
\citet{vinculum} discuss difficult decisions in the design of entity linking systems and benchmarks, which are common sources of linking errors, such as whether to link common entities, how specific the entities should be, which entities to link in case of metonymies, considered entity types and overlapping entities. Many of these decisions motivate our error types in Section \ref{features:error-types}.
\citet{fine-grained-evaluation} manually relabel three benchmarks to evaluate linking systems among dimensions such as the mention’s base form, part of speech and overlap.
\citet{framing-errors} propose an error classification based on the source of the error (e.g. knowledge base errors, dataset errors, annotator errors, etc.). They then manually categorize errors into these classes for a selected set of benchmarks and entity linking systems.
\elevant{} follows this trend and goes one step further by providing a fully automatic classification into fine-grained error categories. Thus, by eliminating the need for expensive human labor, \elevant{} makes the fine-grained evaluation of entity linking errors feasible on a large scale.

\section{Basic Principles}
The core of \elevant{} is a web app that helps users analyze and compare results of various entity linking systems over various benchmarks in great detail.
To this end, the user can add an experiment and evaluate its results using \elevant{}. We define an experiment as a run of a particular linker with particular settings over a particular benchmark.
The pipeline for adding an experiment is as follows:\\[1mm]
(1) add the benchmark (unless it already exists),\\
(2) run an entity linker on that benchmark,\\
(3) automatically evaluate the result in detail.\\[1mm]
The following subsections explain how each of these steps can be executed using \elevant{}. 

\subsection{Adding a Benchmark}\label{basic_principles:adding_benchmarks}
In order to add a benchmark to \elevant{}, it is enough to run a single Python script.
This script converts a given input benchmark into \elevant{}'s internally used article file format. Files in this format contain one JSON object per line which holds information for a single article such as its text, title, ground truth labels or entity predictions.
Additionally, the script annotates ground truth labels with the entities' types (for a fine-grained per-type evaluation) and the entities' names (for presentation purposes).
\elevant{} supports three different benchmark formats: the common NLP Interchange Format (NIF), the IOB format used by \citet{aida-conll} for their AIDA-CoNLL dataset, and a simple JSONL format that only requires information about the benchmark's article texts, the ground truth label spans and the corresponding references to ground truth entities.
Entity references can be from Wikidata, Wikipedia or DBpedia. Entity references from Wikipedia or DBpedia are internally converted to Wikidata.
Several popular entity linking benchmarks are already included in {\elevant} (see Section \ref{features:included_benchmarks}) and can be used out of the box.

\subsection{Running an Entity Linker}\label{basic_principles:running_entity_linkers}
In order to produce entity linking results that can be evaluated with \elevant{}, the user has two options:\\[1mm]
(1) They can feed the output of the entity linker they wish to evaluate into a provided Python script that converts the linking results into \elevant{}'s internal format.
The script supports linking results in NIF, the Ambiverse \cite{aida-conll} output format or a simple JSONL format that only requires information about the predicted entity spans and corresponding entity references.
Like ground truth entity references, references to predicted entities can be from Wikidata, Wikipedia or DBpedia and are converted to Wikidata internally.\\[1mm]
(2) They can implement the entity linker within \elevant{}. The same Python script used for converting entity linking results can then be used to produce new linking results in the required format with the implemented linker.
Several entity linkers are already implemented (detailed in Section \ref{features:included_linkers}) and can be used out of the box.

\subsection{Evaluating Entity Linking Results}
Once the entity linking results are in the required format, they can be evaluated with another Python script.
That script produces output files with the evaluation results.
Using these output files, the results can be instantly viewed in the web app.

\section{Features}

\subsection{Error type classification}\label{features:error-types}

\elevant{} automatically classifies each false positive and false negative into the following three categories and 15 subcategories, to provide detailed information about strengths and weaknesses of a linker.

\paragraph{NER false negatives} are ground truth mentions which the linker did not link to an entity.
They are divided into the following disjunct subcategories:

\begin{itemize}[parsep=1mm,leftmargin=0mm,itemindent=4mm,topsep=1pt]
 \item Lowercased: The first letter in the mention is lower-cased. Linkers that rely on the upper case too much have many errors in this subcategory on benchmarks that contain lower-cased mentions.
 \item Partially included: Not lowercased and a subspan of the mention is linked to an entity. Often a less specific mention is recognized instead of a more specific one, e.g. recognizing ``2022 \mention{World Cup}'' instead of ``\mention{2022 World Cup}''.
 \item Partial overlap: Neither lowercased nor partially included and a span overlapping with the false negative is linked to an entity, e.g. recognizing ``\mention{The Americans}'' instead of ``The \mention{Americans}''.
 \item Other: Remaining undetected mentions.
\end{itemize}

\paragraph{NER false positives} are predicted mentions not labeled in the ground truth. They are further divided into the following disjunct subcategories:
\begin{itemize}[parsep=1mm,leftmargin=0mm,itemindent=4mm,topsep=1pt]
 \item Lowercased: The predicted mention is lower-cased (no named entity) and does not overlap with any ground truth mention.
These are often mentions of abstract entities, which appear in the knowledge base, but are usually not labeled in entity linking benchmarks, for example, \entity{love}{Q316}.
 \item Ground truth entity unknown: The ground truth of the predicted mention is \textit{Unknown}, which means that the true entity is not part of the knowledge base, but an entity from the knowledge base was predicted. Linkers that fail to produce \textit{NIL} predictions have many errors in this subcategory.
 \item Wrong span: The predicted mention overlaps with a ground truth mention that has the same entity, but the spans do not match exactly.
 \item Other: Remaining false detections.
\end{itemize}

\paragraph{Disambiguation errors} are NER true positives that are linked to the wrong entity. They count as false positives \underline{and} false negatives. They are further divided into the following disjunct subcategories:
\begin{itemize}[parsep=1mm,leftmargin=0mm,itemindent=4mm,topsep=1pt]
 \item Demonym: The mention is a demonym (i.e., it is contained in a list of demonyms from Wikidata), such as ``\mention{German}''. Confusions between a country, the people from that country or the language spoken in that country fall into this category.
 \item Metonymy: The mention is a location name (for example, \mention{Berlin}), but the ground truth is not (for example, \entity{government of Germany}{Q159493}), yet the linker wrongly predicted the location.
 \item Partial name: The mention is a part of the ground truth entity's name, e.g., the last name of a person.
 \item Rare: The linker predicted the most popular candidate entity (with candidate sets derived by entity names and Wikipedia hyperlink texts, and popularity measured by the number of Wiki sites about an entity) instead of a less popular one.
 \item Other: Remaining disambiguation errors.
\end{itemize}

\smallskip\noindent
For linkers where we have access to the candidate sets, the following disambiguation error subcategories are reported. They overlap with the previous five subcategories.
\begin{itemize}[parsep=1mm,leftmargin=0mm,itemindent=4mm,topsep=1pt]
 \item Wrong candidates: The ground truth entity is not contained in the candidate set.
 \item Multiple candidates: The ground truth entity is contained in the candidate set, but the wrong candidate was predicted.
\end{itemize}

\subsection{Evaluation per entity type}

\elevant{} assigns a type to each entity and computes precision, recall and F1 score per entity type. Many entity linking benchmarks contain more than the classic person, location and organization entities.
We therefore chose 29 entity types that cover the entities in the included benchmarks well, yet are not too abstract to include many Wikidata entries that are not linked in the benchmarks.
Example types are person, location, organization, languoid, taxon, brand, award, event and chemical entity (for the full list see \elevant{}'s documentation on GitHub).
The types are not restricted to \emph{named} entities, 
but include other types of interest, such as profession, sport and color.
A type $t$ is assigned to an entity $e$, if $t$ and $e$ are connected in a manually corrected Wikidata dump via a property path that starts with an \entity{instance of}{P31} relation and is followed by an arbitrary amount of \entity{subclass of}{P279} relations.

\subsection{Rich visualization}
{\elevant} provides a rich and compact visualization of an entity linker's predictions in comparison to the ground truth labels;
see Figure~\ref{fig:annotations}.
Predictions are shown as highlighted text, while ground truth labels are shown as underlined text.
Both predictions and ground truth labels are color-coded such that true positives, false positives, false negatives and unknown entities can be distinguished at a glance.
On mouse-over, tooltips with additional information about the predicted entity or ground truth entity are shown, such as their Wikidata name and ID.
When the user selects one of the error categories or entity types mentioned above, annotations that fall into the selected category are emphasized.

\subsection{System comparison}
Aside from letting the user compare the evaluation results of different entity linkers in various categories, \elevant{} comes with a feature to compare the predictions of two entity linkers for a selected benchmark side by side.
This allows the user to closely and comfortably examine where differences in the evaluation results of two systems are coming from.

\subsection{Additional Web App Features}
In addition to the prominent features described above, the \elevant{} web app comes with several features that improve overall usability.
Each selectable component such as the benchmark, entity linker, error category or benchmark article has a corresponding URL parameter. The URL is automatically adjusted when a component is selected. This makes sharing the currently inspected results, e.g. the results of a particular linker for a particular error category on a particular benchmark as easy as copying and sharing the current URL.

When evaluating several entity linkers and possibly several versions for each linker, the evaluation results table can quickly become huge. In order to keep the focus on the currently most relevant results, \elevant{} has a filter text field which supports regular expressions. Only experiments whose names match the filter text are displayed in the table.

Our goal was to make the \elevant{} web app as intuitive as possible such that no additional resources would be necessary in order to understand and use it. To this end, the web app itself provides unobtrusive yet easily accessible explanations for its components.
A mouseover button for example gives detailed explanations about the annotations such as the (already intuitive) color code.
Hovering over the table header of an error category opens a tooltip that not only explains the corresponding error category but also gives an example for an entity linker error that falls into this category.
Hovering over precision, recall or F1-score table cells opens a tooltip that shows the total numbers of true positives, false positives, false negatives and ground truth mentions for the corresponding category.

\subsection{Included benchmarks} \label{features:included_benchmarks}

\elevant{} contains the following benchmarks:
\begin{itemize}[parsep=1mm,leftmargin=0mm,itemindent=4mm,topsep=1pt]
	\item AIDA-CoNLL \cite{aida-conll}, a collection of 216 and 231 news articles from the 1990s for validation and testing.
	\item KORE50 \cite{kore50}, 50 difficult, hand-crafted sentences.
	\item MSNBC \cite{msnbc}, 20 news articles from 2007.
	\item MSNBC updated \cite{msnbc-updated}, a version of MSNBC without entities that do no longer exist in Wikipedia.
	\item DBPedia Spotlight \cite{dbpedia-spotlight}, 58 New York Times articles.
\end{itemize}

\subsection{Included linkers} \label{features:included_linkers}

\elevant{} contains pre-computed results of the following entity linkers on the included benchmarks.
\begin{itemize}[parsep=1mm,leftmargin=0mm,itemindent=4mm,topsep=1pt]
	\item TagMe \cite{tagme}
	\item DBpedia Spotlight \cite{dbpedia-linker}
	\item GENRE \cite{genre}
	\item Efficient EL \cite{efficient-el}
	\item Neural EL \cite{neural-el}
	\item Ambiverse \cite{ambiverse-ner} (NER), \cite{aida-conll} (NED)
\end{itemize}

TagMe and DBpedia Spotlight can be run out of the box with \elevant{}. For GENRE, Efficient EL and Neural EL, we provide code with an easy docker setup that yields results in a format supported by \elevant{}.
Furthermore, \elevant{} can process any linking results file which is in NIF, the Ambiverse output format or a simple JSONL format as described in \ref{basic_principles:running_entity_linkers}. These formats are also explained in detail in \elevant{}'s documentation.

Additionally, we include a simple baseline that is based on prior probabilities computed from Wikipedia hyperlinks. The baseline uses the SpaCy \cite{spacy} NER tagger with slight modifications (such as filtering out dates) to detect entity mentions.
All entities with an alias that matches the mention text are considered as candidate entities for a mention.
The aliases of an entity are the anchor texts of incoming intra-Wikipedia hyperlinks to an entity's Wikipedia article, as well as the entity's Wikidata aliases.
From these entity candidates, the entity that has most frequently been linked with the mention text in Wikipedia is predicted.

\subsection{Extendability}
New benchmarks or entity linking results can easily be added to \elevant{} if they are in one of the supported formats using \elevant's conversion scripts as described in Section~\ref{basic_principles:adding_benchmarks} and Section~\ref{basic_principles:running_entity_linkers}.
Additionally, support for other benchmark or entity linking result formats can be added with little effort.
The process for implementing new format readers for benchmarks or entity linking results is explained in \elevant{}'s documentation and existing format readers can be used as templates.

\subsection{Easy knowledge base update}
\elevant{} stores information about entities in several files that are generated from two sources: Wikidata and Wikipedia. The information extracted from Wikidata includes an entity's name, aliases, types and its corresponding Wikipedia URL.
The information extracted from Wikipedia includes intra-Wikipedia link frequencies (how often is a hyperlink's anchor text in Wikipedia linked to a certain Wikipedia article) and Wikipedia redirects which are needed to reliably map Wikipedia entities to Wikidata.
All of these files can either be downloaded from our servers or generated with three simple commands.
The simplicity of the data generation allows for regular updates of the data.

\subsection{Open source}
Our code is open source (Apache License 2.0) and is available on GitHub \footnote{\url{https://github.com/ad-freiburg/elevant/}}.
A Docker setup allows an easy installation and usage.
All links and a web demo are provided at \url{https://elevant.cs.uni-freiburg.de}.

\section{Conclusion}

\elevant{} is a powerful, general-purpose, easy-to-use system for the in-depth evaluation and comparison
of a set of entity linkers on a given set of benchmarks.
Typical evaluations of entity linking systems only provide aggregated figures like precision, recall and the F1 score.
\elevant{} goes beyond this by providing a breakdown of the results by entity type and by error category,
as well as an intuitive visualization of true positives, false negatives, and false positives on the concrete texts.
This can help both practitioners (to understand for which kind of texts a given entity linker is suited)
as well as researchers (to help understand in detail the particular weaknesses of their entity linker and try to improve those).

\bibliographystyle{acl_natbib}
\bibliography{elevant}

\end{document}